%% file: main.tex
\documentclass{article} % For LaTeX2e
\usepackage{iclr2026_conference,times}
\usepackage{amsmath,amssymb,amsfonts,mathtools}
\usepackage{fontawesome5}

\usepackage{algorithm}
\usepackage{algpseudocode} 
\usepackage{xparse}
\usepackage{xcolor}
\usepackage{booktabs, threeparttable}
\usepackage{adjustbox}
\input{math_commands.tex}

\definecolor{citecolor}{HTML}{0071BC}
\definecolor{linkcolor}{HTML}{ED1C24}
\usepackage[pagebackref=false, breaklinks=true, letterpaper=true, citecolor=citecolor, colorlinks, linkcolor=linkcolor, urlcolor=black, bookmarks=false]{hyperref}
\usepackage{cleveref}
\usepackage{url}
\usepackage{multirow}
\usepackage{array} % in preamble
\usepackage{amssymb}% http://ctan.org/pkg/amssymb
\usepackage{pifont}% http://ctan.org/pkg/pifont

\definecolor{lightpurple}{RGB}{242, 242, 255}

\newcommand{\name}{\textsc{DataProphet}}

\iclrfinalcopy % Uncomment for camera-ready version, but NOT for submission.
\NewDocumentCommand{\qixuan}
{ mO{} }{\textcolor{green}{\textsuperscript{\textit{qixuan}}\textsf{\textbf{\small[#1]}}}}

\newcolumntype{x}[1]{>{\centering\arraybackslash}p{#1pt}}
\newcolumntype{y}[1]{>{\raggedright\arraybackslash}p{#1pt}}
\newcolumntype{z}[1]{>{\raggedleft\arraybackslash}p{#1pt}}
\newcommand{\eg}{\emph{e.g}.}
\newcommand{\ie}{\emph{i.e}.}

\usepackage{subcaption}

\usepackage{url}
\newlength\savewidth
\usepackage{booktabs}
\usepackage{xcolor}
\usepackage{todonotes}   % if you want margin notes / inline comments

\title{\includegraphics[scale=0.35]{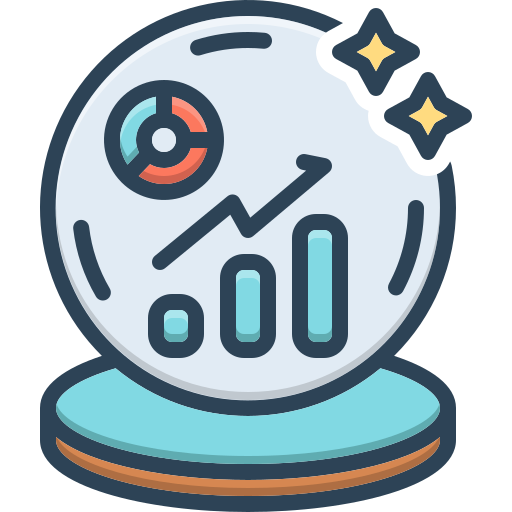} \name: Demystifying Supervision Data Generalization in Multimodal LLMs}
\newcommand{\AffPenn}{\textsuperscript{\(\gamma\)}}  % UPenn
\newcommand{\AffTHU}{\textsuperscript{\(\tau\)}}    % Tsinghua
\newcommand{\AffPU}{\textsuperscript{\(\rho\)}}     % Princeton

\newcommand{\IconWeb}{\faGlobe}        % website / project page
\author{
Xuan Qi\AffPenn\AffTHU\thanks{This work was completed during Xuan Qi's summer research internship at the University of Pennsylvania. Correspond to \texttt{<Xuan Qi: qi-x22@mails.tsinghua.edu.cn>}, \texttt{<Xingyu Fu: xingyufu@princeton.edu>}
}
\quad Luxi He\AffPU
\quad Dan Roth\AffPenn
\quad Xingyu Fu\AffPenn\AffPU \vspace{1mm} \\
{
\AffPenn University of Pennsylvania \quad
\AffTHU Tsinghua University \quad
\AffPU Princeton University
} \vspace{1mm} \\
\texttt{\IconWeb \hspace{0.3em}Website:} \url{https://dataprophet26.github.io/}
\hspace{1em}
\raisebox{-0.4ex}{\includegraphics[height=1em]{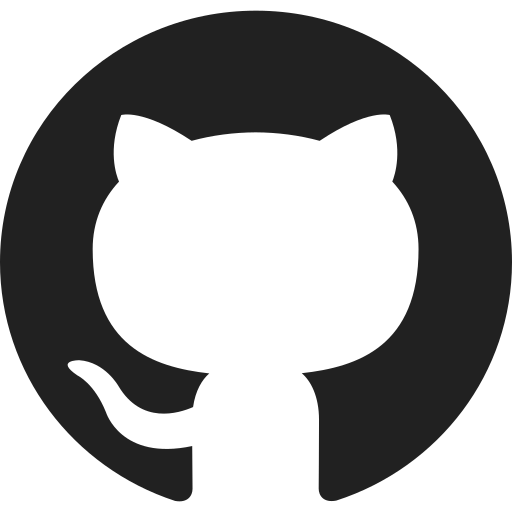}}\hspace{0.3em}\href{https://github.com/DataProphet26/dataprophet}{\texttt{Code}} 
\hspace{0.2cm}
\raisebox{-0.4ex}{\includegraphics[height=1em]{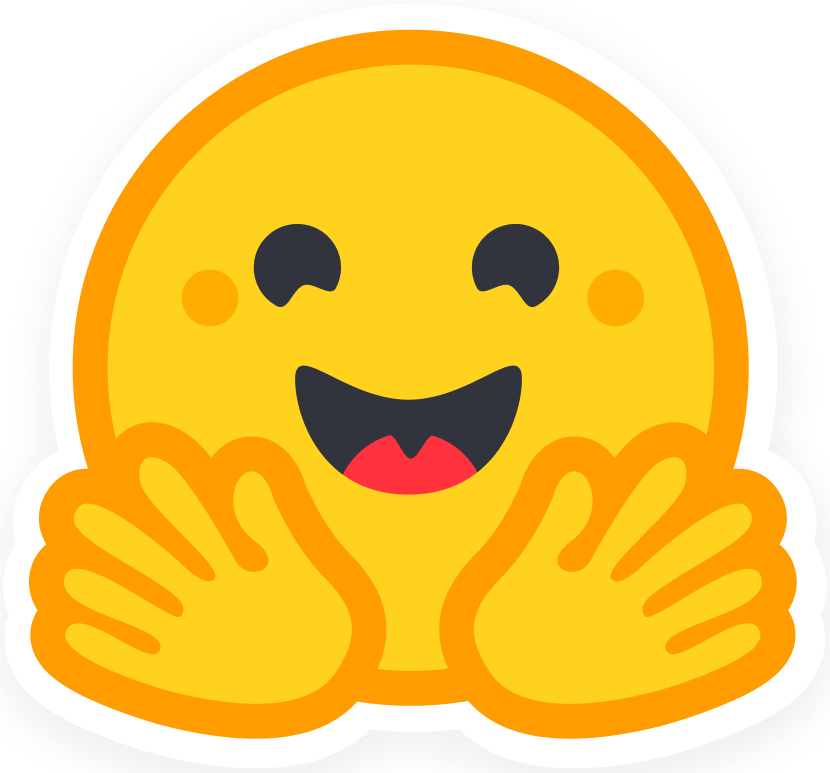}}\hspace{0.3em}\href{https://huggingface.co/datasets/THUQiXuan/DataProphet}{\texttt{Dataset}}
}

\begin{document}

\maketitle

\input{content/0-abstract}
\input{content/1-intro}
% \input{content/3-method}
\input{content/3-cross-task-data-influence}
\input{content/4-metric}
% \input{content/4-experiments}
\input{content/5-data-selection}
\input{content/2-related}
\input{content/6-conclusion}

% \subsubsection*{Author Contributions}
% If you'd like to, you may include  a section for author contributions as is done
% in many journals. This is optional and at the discretion of the authors.

% \subsubsection*{Acknowledgments}
% Use unnumbered third level headings for the acknowledgments. All
% acknowledgments, including those to funding agencies, go at the end of the paper.

\bibliography{iclr2026_conference}
\bibliographystyle{iclr2026_conference}

\appendix
\input{content/appendix}

\end{document}

%% file: math_commands.tex
\usepackage{amsmath,amsfonts,bm}

\def\eqref#1{equation~\ref{#1}}
\def\1{\bm{1}}

\DeclareMathAlphabet{\mathsfit}{\encodingdefault}{\sfdefault}{m}{sl}
\SetMathAlphabet{\mathsfit}{bold}{\encodingdefault}{\sfdefault}{bx}{n}

%% file: content/0-abstract.tex
\begin{abstract}
% \xingyu{Need to extend this analysis to CoT and RL... but we can submit for now}
% \xingyu{Prophet-Index??} 

Conventional wisdom in selecting supervision data for multimodal large language models (MLLMs) is to prioritize datasets that are intuitively similar to the target task (\eg~text-rich \emph{v.s.} vision-centric). However, it remains unclear how reliably such similarity translates into improved performance on the test benchmarks. 
In this paper, we take the first step to study the problem in MLLMs: can we predict a training data's influence on a target benchmark \textit{even before} any training takes place?
% \dr{In this papers we ask whether it is possible to predict the training data's helpfulness on multiple target tasks even \textit{before} any training takes place.}
% \dr{Next, I reversed the order and first talk about the algorithm, and only then the data; seems to me that it flows better. Made a couple other minor changes; please check.}
To answer this question, we first conduct an in-depth analysis using 14 vision-language datasets covering 7 diverse tasks. Our analysis shows that intuitive task
similarity is unreliable in predicting task generalizability, and that transfer depends on the specific dataset rather than the broader task category. 
We propose \name{}, a training-free, simple yet effective metric based on multimodal perplexity, similarity, and data diversity.% We conduct experiments with 14 vision-language datasets covering 7 diverse tasks such as OCR, chart understanding, and spatial reasoning. 
Our experiments demonstrate that the influence rankings for different supervision datasets derived from \name{} is strongly-correlated with rankings based on the actual performance increase after training, with a Kendall’s $\tau$ correlation coefficient of 86.0\%.
Moreover, we show that \name{} can help select better supervision data, achieving up to 6.9\% improvement in average over uniform selection, 1.4\% over SoTA training-based baseline, and 0.2\% higher than oracle experiment performance-based selection. 
%\lucy{@Xuan please check all the highlighted performance numbers in abstract and intro to make sure they're accurate (eg. all the performance increase numbers are aligned with the tables)}
% on 14 benchmarks. 
% By demystifying the generalization achieved given supervision data, we hope \name{} will inspire future research on understanding inter-domain data impacts and creating better open-source data recipes in model training. 
Our code and data will be released. 

\end{abstract}

%% file: content/1-intro.tex
\section{Introduction}

Training data is one of the most important deciding factors for Multimodal Large Language Model (MLLM) performance~\citep{li2023quantity,albalak2024survey,sachdeva2024train,bai2023qwenvl,qwen2vl2024,bai2025qwen2.5vl,zhu2025internvl3,gpt4,team2023gemini}. Given the vast amount of multimodal supervision datasets, how should we effectively utilize them towards certain training targets? Some prior work has explored using high-quality data and similar tasks' data during MLLM supervised fine-tuning stage~\citep{xia2024less,wu2025icons}. However, these methods still require some training and mainly improve supervision performance on target tasks by removing irrelevant, redundant, or low-quality data from the training mixture. 
% \lucy{but icons method itself does not focus on removing irrelevant data?} 
In this work, we further demystify how training data affects a test benchmark's performance. We pose the following research question in an MLLM setting: \begin{center}
\colorbox{lightpurple}{\parbox{0.95\linewidth}{
Given a training dataset $D_i$, can we predict the influence (measured by relative performance change) on a target benchmark $T_j$ \textit{even before} any training takes place? 
}}
\end{center}

\begin{figure}[t!]
\centering
\includegraphics[width=1\linewidth]{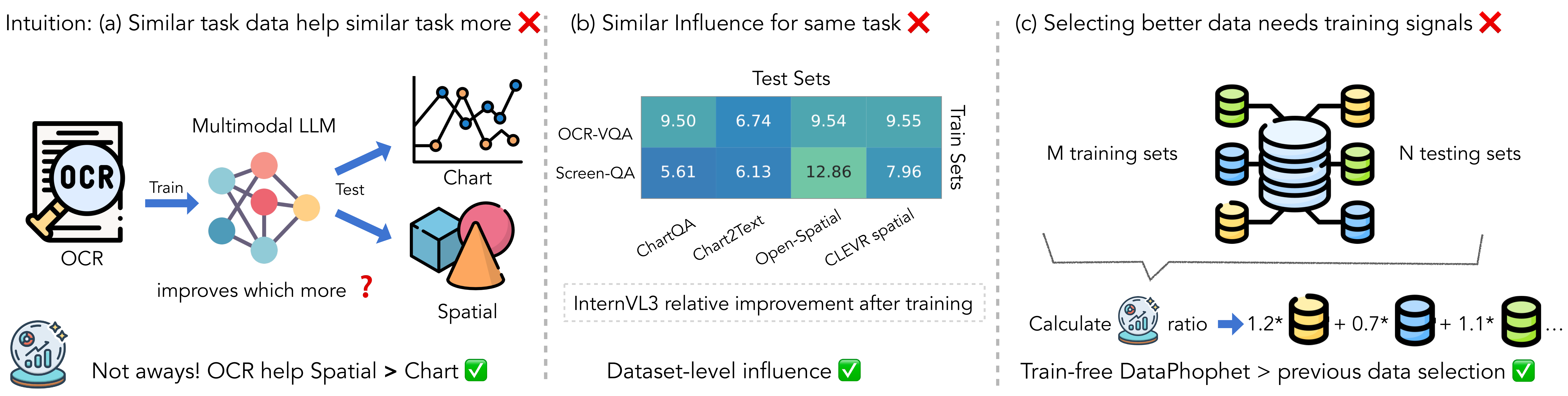}
\caption{\textbf{Three major takeaways in \name}: (a) Surprisingly, human intuition about similarity between training data and test data turns out to be a poor indicator of model performance influence. In contrast, our designed \name{} metric more reliably predicts the influence of training data on test benchmarks; (b) the impact of multimodal supervision is decided by specific individual datasets, rather than by broad task categories: datasets from the same task category do not necessarily help each other the most, and do not share similar influence on the same target benchmark. Here, OCR-VQA~\citep{mishra2019ocr} and Screen-QA~\citep{hsiao2022screenqa} are both OCR data, ChartQA~\citep{masry2022chartqa} and Chart2Text~\citep{kantharaj2022chart} are both chart question answering benchmarks, and Open-Spatial~\cite{cheng2024spatialrgpt} and CLEVR spatial~\citep{johnson2017clevr} are both spatial reasoning benchmarks; (c) \name{} provides an effective approach for training-free data selection under fixed compute budgets (\ie fixed number of total training samples). We compute influence of each supervision dataset based on a combined set of testing benchmarks and select data according to the influence ratio, yielding consistent improvements across 14 tasks, with average gains of +3.4\% and +6.9\% under real and synthetic data settings.}
\label{fig:teaser}
\vspace{-2em}
\end{figure}

To answer this question, we first conduct a comprehensive experiment utilizing 14 vision-language fine-tuning datasets covering 7 diverse tasks (2 for each task): OCR, chart understanding, spatial reasoning, counting, knowledge-based QA, document understanding, and map understanding. We surprisingly find that different training datasets' influence on a target benchmark may contradict human intuition. For example, humans may intuitively assume that training the model on OCR task data will help its performance on chart tasks more than spatial reasoning tasks, since OCR and chart both require extracting text and numbers in an image. However, experiment results as in \Cref{fig:teaser}(a) and (b) indicate the reverse: training on OCR improves spatial reasoning tasks more than chart tasks. This counter-intuitive finding suggests that model's generalization across different tasks depends on more than surface-level similarity. We also find that performance improvement are not decided by task category but are dependent on individual datasets: datasets from the same task category do not necessarily help each other the most, and do not share same influence on the same target benchmark.

Based on our analysis, we extract factors that \emph{truly matter} for predicting such influence. We introduce \name, an extremely simple yet effective, training-free metric which uses text and visual similarity, multimodal perplexity, and data diversity to predict the data influence (\Cref{eq:metric-final}). We evaluate the ranking of 14 training datasets' influence compared to ranking using real training effect, measured by the benchmark's relative improvement over baseline. Our experiments show strong correlation between the two rankings (86.0\% Kendall’s $\tau$), demonstrating the usefulness of \name{} in predicting data influence rankings in scale. Notably, the most important factors for prediction are multimodal perplexity and visual similarity, increasing 37.3\% and 23.5\% of Kendall’s $\tau$ respectively.

% focus on is human-annotated high-quality training data that is in same-distribution as test tasks. 
% Also, these methods all require training, as using gradient-based solutions~\citep{wu2025icons,xia2024less}. What we look for is a training-free approach to do the task. What's more, these methods could only rank training data's influence based on test data, can we do it also in the opposite direction, to rank different test data's improvement given one training data? Will text matter more or vision matter more for multimodal tasks?

In addition to predicting source data usefulness, we also show that given target benchmarks, \name{} can be applied to curate good instruction-tuning data under fixed compute and data budget. We conduct experiments under two different settings: selecting instruction-tuning data from (1) real data pool and (2) synthetic data pool. For real data, \name{} performs 3.4\% better than uniform selection and is 1.4\% better than ICONS~\citep{wu2025icons}, a SoTA selection requiring training. We observe even larger performance gain in synthetic data selection, reaching almost 6.9\% performance increase compared to the uniform selection baseline and 1.2\% improvement over ICONS. These show promise for \name{} as a light-weight yet effective measure for choosing supervision data for MLLMs.

Overall, this work performs a deep analysis of different source datasets' influence on target benchmarks' performance, with counter-intuitive findings. Our analysis motivates \name{}, a simple training-free yet effective metric to predict the usefulness of different training datasets on the target. We further demonstrate the promising effect of \name{} to guide MLLM instruction-tuning data selection that outperforms SoTA training-based selection. 

%  For real data case, \name{} could perform 1\% better than training-required gradient-based data selection methods, and 3.4\% better than random sampling, using same amount of compute. For synthetic data case, \name{} could perform 1\% better than training-required data selection method, and 7\% better than random sampling.
% We further analyze about kind of synthetic data gets selected out, \eg whether GPT or Gemini provides better quality, and conduct a human analysis on the difference between human-selected high quality and low quality data versus model selected ones.

% The major takeaways of the paper are as follows: (1)similar task's training data do not necessarily help similar test tasks more as humans would imagine \eg text-rich VQA v.s. vision-centric VQA; (2) multimodal supervision data influence is dependent on specific datasets. Different datasets within the same task or domain can have different effects on the same target benchmark;  (3) under fixed compute,  \name{} selects better supervision data than training-based methods.

% \begin{figure}[t]
% \centering
% \includegraphics[width=1\linewidth,trim={1.0cm 1.5cm 0cm 1.3cm},clip]{figures/relative.pdf}
% \caption{\textbf{Relative improvement} (Eq.~\ref{eq:rel-impr}) for every train$\to$test pair. Rows: training data; columns: test sets. Numbers are percentage gains over the baseline (InternVL3-2B) on each test set. \lucy{Add some note-worthy observations here.}}
% \label{fig:rel-heatmap}
% \end{figure}

%% file: content/3-cross-task-data-influence.tex
\section{Data Influence Analysis}
\label{sec:cross_task_influence}
In this section, we demonstrate how we quantitatively analyze data influence across multiple visual question answering (VQA) tasks. 
% \paragraph{Quantifying cross-task data influence.}
% We would like to study how much does training on one dataset (source) improve performance on a different dataset (target) across diverse multimodal tasks. 
% To that end, we measure \emph{cross-task data influence} 
Specifically, we first select a base Multimodal Large Language Model (MLLM): InternVL3~\citep{zhu2025internvl3}, and 14 diverse VQA datasets that comes with both train (source) and test (target) sets. Then, we conduct supervised fine-tuning on our base model with each source dataset individually, and evaluate relative performance gains on all target sets. The goal is to demystify data influence from real experiment results. We will expand this section by introducing the experiment setup, followed with results and observations.

\subsection{Analysis Setups}
\paragraph{Benchmark and Task Selection.}
We try to include as diverse benchmarks as possible and mainly select three types of tasks: text-rich tasks such as OCR, chart, and document; general VQA tasks; and vision-centric tasks such as spatial reasoning, counting, and map understanding. This ends in 14 vision–language datasets spanning seven task families (2 datasets for each task), as detailed below:
\begin{itemize}
  \item \textbf{OCR:} OCR-VQA ($\sim$1002K QA pairs)~\citep{mishra2019ocr}; ScreenQA ($\sim$86K QA pairs)~\citep{hsiao2022screenqa}. This task requires extracting specific text from text-rich images and reasoning about its context.
  
  \item \textbf{Chart understanding:} ChartQA ($\sim$33K QA pairs)~\citep{masry2022chartqa}; Chart2Text ($\sim$44K QA pairs)~\citep{kantharaj2022chart}. Models must interpret chart data, identify trends, and answer questions or generate textual descriptions of charts.
  
  \item \textbf{Document understanding:} DocVQA ($\sim$50K QA pairs)~\citep{Mathew_2021_WACV}; Sujet-Finance-QA ($\sim$107K QA pairs)~\citep{sujetfinanceqa2024}. This task requires extracting relevant information from structured and unstructured text in documents to answer questions.
  
  \item \textbf{General VQA:} A-OKVQA ($\sim$24.9K QA pairs)~\citep{schwenk2022aokvqa}; VC-GVQA extracted from Visual-CoT ($\sim$438K QA pairs)~\citep{shao2024visualcot}. This task focuses on general domain question-answering over natural images.
  
  \item \textbf{Spatial reasoning:} Open-Spatial ($\sim$8.7M QA pairs)~\citep{cheng2024spatialrgpt}; CLEVR-Relation ($\sim$160K QA pairs)~\citep{johnson2017clevr}. This task tests model's ability to understand spatial relationships between objects, and perform relational reasoning.
  
  \item \textbf{Counting:} CLEVR-Counting ($\sim$160K QA pairs)~\citep{johnson2017clevr}; TallyQA ($\sim$287K QA pairs)~\citep{acharya2019tallyqa}. This task requires accurately counting objects in images or identifying quantities based on visual cues.
  
  \item \textbf{Map reasoning:} GeomVerse ($\sim$23K QA pairs)~\citep{kazemi2023geomverse}; MapQA ($\sim$800K QA pairs)~\citep{chang2022mapqa}. This task focuses on understanding maps, spatial navigation, and reasoning about geographic or structural layouts.
\end{itemize}
Unless otherwise noted, we use the officially released splits or collections. For CLEVR-based subsets ({CLEVR-Relation} and {CLEVR-Counting}), we follow the question-family filters described in~\cite{johnson2017clevr} and train/evaluate on their corresponding target sets.

\begin{figure}[t]
\centering
\includegraphics[width=1\linewidth,trim={0.0cm 0cm 0cm 1.3cm},clip]{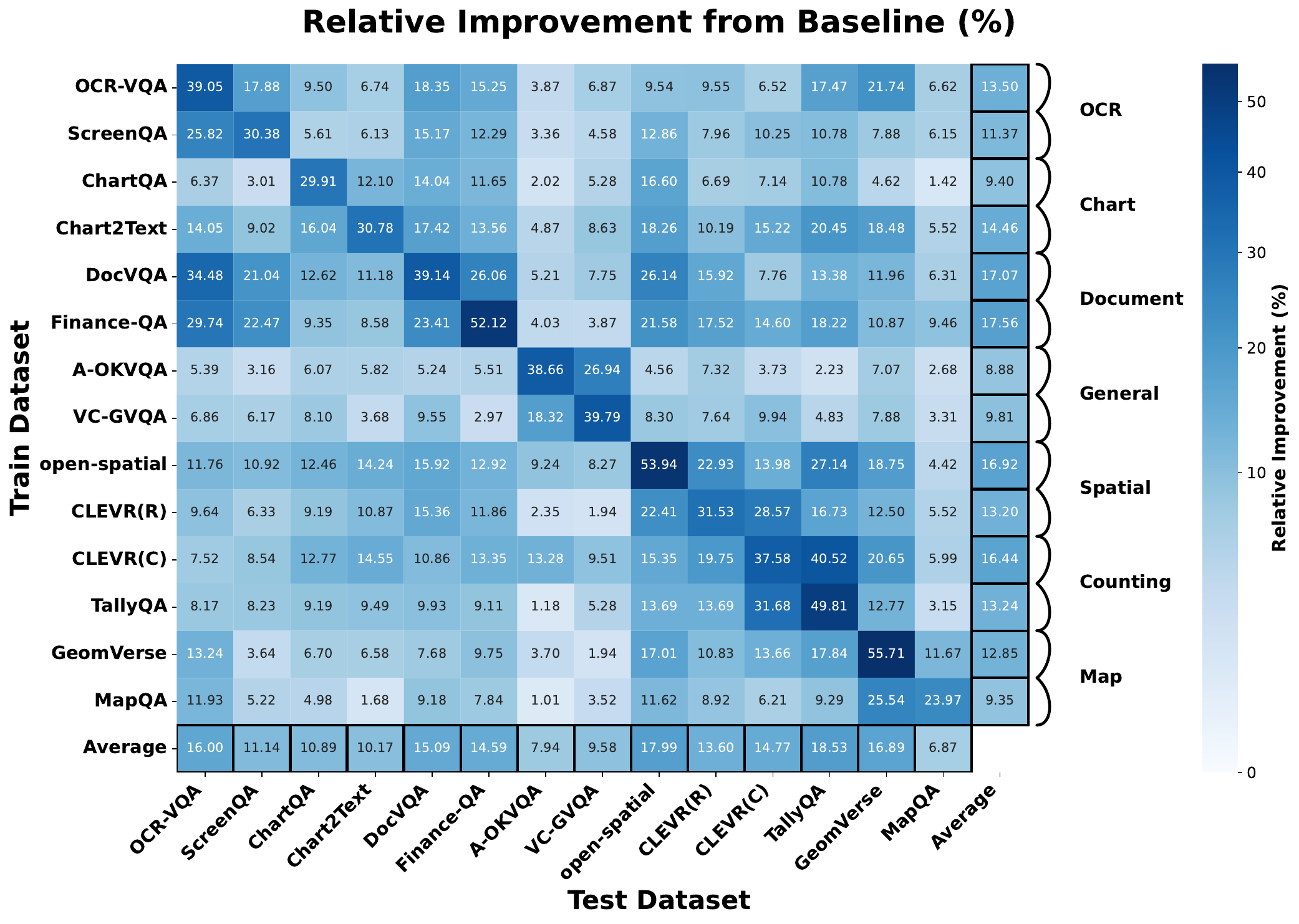}
\caption{\textbf{Data influence analysis under fixed compute.}
We conduct supervised fine-tuning on base model (InternVL 3~\citep{zhu2025internvl3}) with each training (source) dataset individually, and evaluate \emph{relative performance gains} on all test (target) sets. See details in \Cref{sec:cross_task_influence}.
We highlight several observations: (1) Data influence is not symmetric. (2) Data sources from the same task type do not always influence each other most. E.g., OCR-VQA improves ScreenQA (OCR task) by 17.88\% vs. GeomVerse (map understanding task) by 21.74\% (3)  Data influence is not decided by task type, but by the individual dataset. E.g., text-rich tasks (ScreenQA) can influence vision-centric ones (Open-Spatial with gains of 12.86\%) more than text-rich ones (ChartQA with gains of 5.61\%).
}
\label{fig:rel-heatmap}
\vspace{-2em}
\end{figure}

% \paragraph{Protocol.}
% Let \(A_t\) denote the pretrained baseline model's task-specific score on test set \(t\) (e.g., accuracy), and \(A_t^s\) the score after fine-tuning on source dataset \(s\).
% We quantify the source-to-target influence via \emph{relative improvement} $\Delta_{s\to t} \;=\; \frac{A_t^s - A_t}{A_t}\,.$
% \begin{equation}
% \label{eq:rel-impr}
% \Delta_{s\to t} \;=\; \frac{A_t^s - A_t}{A_t}\,.
% \end{equation}

\paragraph{Experiment Setup.}
For each benchmark \(\mathcal{D}\), we uniformly sample \(20\mathrm{k}\) examples from their training splits for supervised fine-tuning, and \(1\mathrm{k}\) from test splits for evaluation. We denote the training (source) data as $s$, and testing (target) data as $t$.
All experiments are conducted with a fixed compute setting -- same number of training data (20K), same base model, and same hyper-parameter choices (optimizer, scheduler, epoch count, and batch size).
The base model we use here is \textbf{InternVL3-2B}\footnote{\url{https://huggingface.co/OpenGVLab/InternVL3-2B}}~\citep{zhu2025internvl3}.  The implementation details can be found in~\autoref{appen:implementation-details}.
% All runs fine-tune the base model for one epoch using full-parameter supervised fine-tuning (Full SFT) implemented in LLaMA-Factory~\citep{zheng2024llamafactory}. All evaluations follow the benchmarks' task-appropriate metrics.
% \lucy{For detailed hyperparameter setup, please see XXX (add in appendix)}

\paragraph{Relative Performance Gain.}
Here, we introduce how we measure the data influence in real experiment results.
Let \(A_t\) denote the base model's performance score on test set \(t\) (e.g., accuracy), and \(A_t^s\) the score after fine-tuning on source dataset \(s\).
We quantify the source-to-target influence via \emph{relative improvement} $\Delta_{s\to t} \;=\; \frac{A_t^s - A_t}{A_t}\,.$ Similarly, define $\Delta_{t\to s}=\frac{A_s^{\,t}-A_s}{A_s}$, where $A_s$ is the base score on $s$ and $A_s^{\,t}$ is the score on $s$ after fine-tuning on $t$. %\lucy{Add definition for the other-way delta. We use it multiple times afterwards but it was not defined.}

\subsection{Analysis Results}
We finetune our base model with each source dataset individually, and evaluate \textit{relative performance gains} on all target sets, to demystify data influence through controlled real experiment results. We report the full experiment results as in Figure~\ref{fig:rel-heatmap}. We highlight the below observations:

\textbf{Data influence is not symmetric}. The diagonal gains are still the largest, which align with our intuition that data coming from the same distribution is still the most influential. However, we find that the data influence is not symmetric, i.e. $\Delta_{s \to t} \;\neq\; \Delta_{t \to s}.$ For instance, the influence of Open-Spatial on DocVQA (train on Open-Spatial and test on DocVQA) is 15.92\% while the reverse is high as 26.14\%. 

\textbf{Data sources from the same task type do not always influence each other most}. Data influence is not decided by task type, but by the individual dataset. For instance, after training on OCR-VQA data, the relative performance gain achieved on ScreenQA (which is also OCR) is 17.88\%, lower than that achieved on GeomVerse (which is on map understanding) 21.74\%.

\textbf{Text-rich tasks can influence vision-centric ones more than text-rich ones}. As demonstrated in \Cref{fig:rel-heatmap,fig:teaser},  after training on OCR data (ScreenQA), the relative performance gain model achieves on chart tasks -- 5.61\% for ChartQA and 6.13\% for Chart2Text -- is lower than that on spatial reasoning tasks (12.86\% for Open-Spatial and 7.96\% for CLEVR-Relation).
% \lucy{@Xingyu let's discuss this later? Do we have results for each of the bulletpoints here? For presentation format I also think we should do it as paragraphs instead of bulletpoints.}

% \textbf{ vs. cross-domain transfer.} \textit{[To be completed: compare OCR/Doc \(\rightarrow\) text-rich targets vs.\ vision-centric targets; note any asymmetric influences.]}

% \textbf{Dataset-specific idiosyncrasies.} \textit{[To be completed: contrast, e.g., \textsc{OCR-VQA} vs.\ \textsc{ScreenQA} as sources; \textsc{ChartQA} vs.\ \textsc{Chart2Text}; \textsc{CLEVR} subsets.]}

% \textbf{Negative transfer patterns.} \textit{[To be completed: identify source--target pairs with consistent drops and hypothesize confounding factors.]}

% \textbf{Scaling with target difficulty.} \textit{[To be completed: stratify targets by baseline difficulty (accuracy/PPL) and relate to average \(\Delta_{s\to t}\).]}

% \textbf{Compute sensitivity.} \textit{[To be completed: discuss robustness of trends under alternative budgets (e.g., \(10\mathrm{k}\) / \(40\mathrm{k}\)).]}

%% file: content/4-metric.tex
\section{\name{}: Demystify Data Influence}
\label{sec:metric}
\Cref{sec:cross_task_influence} shows that intuitive similarity is unreliable in predicting task generalizability. In this section, we investigate what the true underlying factors for predicting data influence could be. Our goal is to identify a reliable data influence predictor based on these factors.
% In this section, we seek to understand the underlying factors that drive the real data influence results that we collected in \Cref{sec:cross_task_influence}.
We begin with defining a evaluation for such dataset influence prediction metrics (\S\ref{subsec:metric-tau-eval}). We then introduce details of our metric (\S\ref{subsec:metric-definition}). Our metric is motivated by some heuristic functions that people have studied in Large Language Models (LLMs) research, such as data perplexity, task difficulty, and answer length ~\citep{marion2023less, xie2023dsir, wang2024diversitySubsetSelection},\, and we try to study their impact in multimodal settings. Building on these heuristics, we identify a training-free metric for MLLMs that reliably predicts the actual data influence results, not only for $\Delta_{s \to t}$ but also for $\Delta_{t \to s}$. We conclude the section with results of our metric and analysis including different components' effect, and ineffective but heuristic ones that we explored(\S\ref{subsec:metric-results}). 

\subsection{Predictor Evaluation Setup}
\label{subsec:metric-tau-eval}
 Considering the \emph{asymmetric nature} of data influence discussed previously, a good data influence predictor should be one that (1) produces valuable insights on how different datasets compare in contributing to some target task, and (2) offers understanding on how one dataset could affect various downstream targets differently. The ground-truth for these can be approximated by the actual relative performance gain after finetuning and evaluation, following the setup in~\Cref{sec:cross_task_influence}.
%  experiments, based upon the same 14 source and 14 target datasets. 
% Considering the \emph{asymmetric nature} of data influence, we evaluate in a \emph{two-way} format.
% . We use the same model and apply the finetuning and performance evaluation as before. 

\paragraph{Two-Way Evaluation protocol.}
For each pair of (source dataset \(s\) , target data \(t\)), we compute the influence prediction score between them using the data influence predictor. Since we have 14 datasets, we get 256 (14$\times$14) scores in total. Given one target \(t\) dataset, we rank all 14 source datasets using the influence prediction score (estimate of training data's helpfulness), and compare that with the real ranking using the relative improvement scores as in \Cref{fig:rel-heatmap} (true helpfulness). To quantify the correlation between two rankings, we deploy Kendall’s $\tau$. We denote the correlation as $\tau_{t}$ for measuring $\Delta_{s \to t}$, and the average across all target datasets is $\overline{\tau}_{\text{Tgt}}$.

% We take the average of correlation coefficients across all target / source datasets, and report the two types of ranking correlations above, denoted as $\overline{\tau}_{\text{Tgt}}$ and $\overline{\tau}_{\text{Src}}$ respectively.

Conversely, given one source \(s\) dataset, we rank all target datasets based on the influence score between them. Then we similarly compare it with the real rankings as in \Cref{fig:rel-heatmap}. We denote the Kendall’s $\tau$ correlation as  $\tau_{s}$ in this setting for measuring $\Delta_{t \to s}$, and the average across all source datasets is $\overline{\tau}_{\text{Src}}$.
We use $$\overline{\tau} = \frac{\overline{\tau}_{\text{Tgt}}+\overline{\tau}_{\text{Src}}}{2}$$ as the final evaluation metric for influence predictor. Here, the two-way protocol allows for a more comprehensive evaluation of the data influence predictor itself. However, when using the influence predictor to select better training data as in \Cref{sec:data-selection}, we only need to select by $\overline{\tau}_{\text{Tgt}}$. 

% For each target dataset \(t\):
% (i) we obtain the \emph{observed} improvements \(\{\Delta_{s\to t}\}_s\) as described in~\autoref{sec:cross_task_influence};
% (ii) we compute the {predicted} scores of our metric \name{} \(\{\mathcal{M}(s\!\to\!t)\}_s\);
% (iii) we compare the two induced rankings via Kendall’s \(\tau\) and report both per-target \(\tau(t)\) and the macro-average \(\overline{\tau}\) across all 14 targets.
% % \paragraph{Evaluation protocol.}
% For each target dataset \(t\), we compute 
% Kendall’s $\tau$ between the ranking of source datasets by observed relative improvements $\{\Delta_{s\to t}\}_s$ and the ranking induced by the training-free \name{} scores $\{\mathcal{M}(s\!\to\!t)\}_s$. 
% Higher correlation coefficients indicate more alignment between the source dataset helpfulness ranking predicted by \name{} and that induced by the actual performance gain. 
% We take the average of correlation coefficients across all datasets and report $\overline{\tau}_{\text{tgt}}$ as the final metric.
% ------------------------------------------------------

% \begin{figure}[t]
% \centering
% \includegraphics[width=\linewidth]{figures/result1.pdf}
% \caption{\textbf{Predictive accuracy of \name{} (Kendall’s \(\tau\)).}
% For each target (column), we compute the correlation between the source ranking induced by observed relative improvements \(\{\Delta_{s\to t}\}_s\) and the ranking induced by the predicted scores \(\{\mathcal{M}(s\!\to\!t)\}_s\).
% The rightmost column reports the macro-average across all 14 targets.
% Higher is better.}
% \label{fig:kendall-matrix}
% \end{figure}

\subsection{The \name{} Metric}
\label{subsec:metric-definition}
There are three major components in our prediction metric, building on previous work in LLM-setting and incorporating MLLM-specific considerations:
% We seek to understand the underlying factors that drive cross-task transfers and develop a training-free metric that can predict how much a source dataset \(\mathcal{D}\) will help on a target dataset \(\mathcal{T}\). Motivated by the patterns identified in \Cref{sec:cross_task_influence}, we develop a predictive metric that takes into account the following aspects \lucy{Add some motivation in each paragraph based on previous section}

\textbf{Multimodal perplexity.} 
Previous studies~\citep{marion2023less,thrush2024improving} have explored to use perplexity as data quality estimator in language model pretraining, ranking and pruning pretraining corpora accordingly to achieve better model performance. Especially \cite{thrush2024improving} points out the perplexity-benchmark correlations in LLMs. 
Inspired by them, we examine if multimodal perplexity is the best factor for this predictor. We hypothesize that intrinsically more challenging source data can provide greater value for enhancing model capability. We define multimodal perplexity as below:\\
Let \(A=(a_1,\dots,a_T)\) be the answer tokens, and \(p_\theta\) the base MLLM. We define multimodal perplexity on a dataset \(\mathcal{D} \) as
\begin{equation}
\label{eq:ppl}
\text{PPL}(\mathcal{D}) \;=\;
\exp\!\Big(-\,\mathbb{E}_{(I,Q,A)\sim \mathcal{D}}\,\mathbb{E}_{t\sim [T]}\,
\log p_\theta\!\big(a_t \,\big|\, I,\,\tau_Q(Q),\, a_{<t}\big)\Big),
\end{equation}
% following standard practice in language modeling~\citep{jelinek1997statistical}. This factor rewards inherently challenging source data (high $\text{PPL}(\mathcal{D})$), and we further balance it with the intrinsic difficulty of the target dataset ($\text{PPL}(\mathcal{T})$). 
We evaluate the data influence rankings using only multi-modal perplexity as predictor, and find that the resulting $\overline{\tau}_{\text{Tgt}}$ is 0.274, which is not high enough, indicating that additional components are needed for a better predictor.

\textbf{Cross-dataset Similarity.}
If the source dataset contains images and text similar to the target in the model's embedding space, then heuristically it should more closely resemble learning in the target domain's distribution. 
We measure alignment along \emph{questions}, \emph{answers}, and \emph{images} using the output of the (frozen) encoders of the base MLLM, and calculate the similarity for each of them separately. 
For each field of question, answer and image, we encode samples with appropriate templates, then do the \(\ell_2\)-normalization for the embeddings, and take the \emph{expected cosine similarity} between independently drawn items from \(\mathcal{D}\) and \(\mathcal{T}\). We denote these expectations by \textit{QSim}, \textit{ASim}, and \textit{ISim}. 
This factor rewards sources whose question and answer formats and visual layouts resemble the target.
Interesting, we first tried to encode question and answer together as \textit{QASim}, but it showed worse than separate version by 6\% on $\overline{\tau}$ score. We evaluate the data influence rankings using only multi-modal perplexity times similarity as predictor, and find that the resulting $\overline{\tau}_{\text{Tgt}}$ is 0.658, still not satisfactory.
% \xingyu{@Xuan  Do we have this??? We evaluate the data influence rankings using only multi-modal perplexity times similarity as predictor, and find that the resulting $\overline{\tau}_{\text{Tgt}}$ is XXX, still not satisfactory.}

\textbf{Source dataset question diversity.}
Heuristically, if a source dataset contains a diverse set of questions, fine-tuning on it may encourage the model to acquire more generalizable skills, as previous papers have pointed out in LLMs~\citep{wang2024diversitySubsetSelection}. To quantify this diversity, we construct an embedding representation $\mathbf{z}_u$ for each source data's example question \(u\). We then cluster the embeddings into $K$ groups by certain clustering algorithms such as K-means, and compute the silhouette coefficient~\citep{rousseeuw1987silhouettes} given by
\begin{equation}
\label{eq:silhouette}
\mathrm{Sil} \;=\; \mathbb{E}_{u\sim\mathcal{U}}\!\left[\frac{b(u)-a(u)}{\max\{a(u),\,b(u)\}}\right],
\end{equation}
where \(a(u)\) is the mean intra-cluster distance and \(b(u)\) is the minimum mean distance from \(u\) to any other cluster (both measured in the embedding space). We also quantify cluster \emph{balance} via the normalized entropy
\(\mathrm{H} = -(\log K)^{-1}\sum_{k=1}^{K} \pi_k \log \pi_k\) with \(\pi_k\) the empirical cluster proportion. The final diversity is thus \(\mathrm{Sil}+\mathrm{H}\), which larger value indicating greater coverage and more balanced, well-separated clusters.
In experiments, we construct \(\mathbf{z}_u\) by averaging field embeddings and apply \(k\)-means algorithm for clustering with $K=10.$

\paragraph{\name{} Metric.}  Combining the three components above, we introduce our \emph{training-free}, simple yet effective, data influence predictor \name{}
\begin{equation}
\label{eq:metric-final}
\mathcal{M}(s\!\to\!t)
\;=\;
\frac{\text{QSim}\cdot \text{ASim}\cdot \text{ISim}\cdot \text{PPL}(s)\cdot \big(\mathrm{Sil}+\mathrm{H}\big)}
     {\text{PPL}(t)},
\end{equation}
where s is the training (source) data and t is the testing (target) data. 
\name{} implies that successful transfer requires simultaneous alignment in text, vision, difficulty to base model, and question coverage. This thus motivates the product form in our metric, which down-weights the score when any single factor is weak. We use finite-sample estimates for all expectations; when subsampling is required, we select uniformly from the full set. 
% \medskip
% \noindent\textit{Remarks.} (a) \(\mathcal{M}\) is inherently \emph{directional}: \(\mathcal{M}(\mathcal{D}_s\!\to\!\mathcal{D}_t)\neq \mathcal{M}(\mathcal{D}_t\!\to\!\mathcal{D}_s)\).
% (b) Alternatives such as additive or log-linear combinations were explored and showed slightly weaker average rank accuracy (Appendix).

% ------------------------------------------------------

\subsection{Predictor Results and Analysis}
\label{subsec:metric-results}
\input{tables/tao}
% \paragraph{High rank accuracy across targets.}
% \name{} tracks the empirical transfer rankings with high fidelity: the \textbf{average} Kendall correlation is \textbf{0.863}. ~\autoref{tab:kendall-by-category} summarizes the per-target correlations, showing consistently strong alignment between our metric's predictions and actual fine-tuning outcomes.

\paragraph{\name{} is simple but effective.} While heuristically based and training-free, our results in \Cref{tab:kendall-by-category} suggests that \name{} can successfully predict the data reflectance as in real training results. 
The $\overline{\tau}_{\text{tgt}}$ is consistantly high across all target datasets and task types, with an average value of 0.863. And the $\overline{\tau}_{\text{src}}$ also reaches 0.857 across all source datasets, indicating the effectiveness of our metric.
% \xingyu{The $\overline{\tau}_{\text{src}}$ also reaches XXX all source datasets,!} 
% This shows that \name{} is a useful metric for predicting the usefulness ranking of source datasets on given target task. 

% Furthermore, our results are robust to variations in the experiment setup. $\overline{\tau}_{\text{tgt}}$ remains high when we halve or double the SFT data budget to 10k and 40k respectively or vary \(K\in\{16,32,64\}\). The average changes with setup variations are within \(\pm 0.02\) Kendall’s \(\tau\) (see Appendix for details and confidence intervals \lucy{@Xuan add to appendix}).

% \subsection{Analysis and Ablations}
% \label{subsec:metric-analysis}

\paragraph{Importance of individual components.} We further investigate the contribution of each component to our metric. Removing each factor in \Cref{eq:metric-final} degrades rank accuracy (Table~\ref{tab:ablation-components}). The most pronounced drop occurs when \emph{multimodal perplexity} is removed (\(0.863\to 0.491\)), highlighting the value of modeling source headroom and target difficulty. \emph{Image similarity} (\(0.627\)) and \emph{diversity} (\(0.658\)) are also crucial components, suggesting that visual alignment and coverage are essential for transfer. Textual similarities from questions and answers are useful, but their contributions are more marginal compared to the other stronger signals.

\begin{table}[t]
  \centering
  \begin{threeparttable}
    \setlength{\tabcolsep}{8pt}
    \begin{adjustbox}{max width=\textwidth}
    \begin{tabular}{lc}
      \toprule
      \textbf{Variant} & \textbf{Kendall’s \(\tau\) (Avg)} \\
      \midrule
      Full \name{}                               & \textbf{0.860} \\
      w/o Answer Similarity                      & 0.810 \\
      w/o Question Similarity                    & 0.778 \\
      w/o Diversity (Silhouette \& Entropy)      & 0.659 \\
      w/o Image Similarity                       & 0.625 \\
      w/o Perplexity                             & 0.487 \\
      \bottomrule
    \end{tabular}
    \end{adjustbox}
    \caption{\textbf{Ablation of \name{} components} Each row indicates removing one factor from Equation~\eqref{eq:metric-final} and recompute Kendall’s $\tau$. Different components have varying effects on the correlation coefficient.
    % \lucy{Need to change it from average tau to both direction tau's.\xingyu{if it's not huge difference we can change it after submission...}}
    }
    \label{tab:ablation-components}
  \end{threeparttable}
\end{table}

% The findings here mirror our observations in \Cref{sec:cross_task_influence}:
% (1) sources that are \emph{intrinsically harder} for the baseline (higher \(\text{PPL}(\mathcal{D}_s)\)) tend to be stronger teachers; and
% (2) sources that are \emph{visually aligned} and \emph{diverse} transfer more broadly, consistent with the large contributions of ISim and the diversity term. \lucy{@Xingyu could you add connection to end of section 2 here? The original writing says this mirrored previous observation but I don't think there is very strong connection here?}

\paragraph{Additional trials.} We conduct multiple additional heuristic-based trials but they all fail to show improvement: question difficulty, model's familiarity with the image, model's familiarity with the question, and answer length. For example, after incorporating answer length, the $\overline{\tau}$ decreased by $0.15$. Only the three major factors in our \name{} metric remains effective.

% \xingyu{@Xuan add numbers if you have any, e.g. the tau changes from xxx to xxx after adding xxx}

% \paragraph{Ranking test-set with train}
% Second, fixing each source $s$ (rows), we compute Kendall’s $\tau$ between $\{\Delta_{s\to t}\}_t$ and $\{\mathcal{M}(s\!\to\!t)\}_t$, and average over sources to obtain $\overline{\tau}_{\text{src}}$. Our final summary is the mean of the two macro-averages, $\overline{\tau}=\tfrac{1}{2}(\overline{\tau}_{\text{tgt}}+\overline{\tau}_{\text{src}})$.

% We first fix each target and average across targets to obtain $\overline{\tau}_{\text{tgt}}$. We then fix each source and average across targets to obtain $\overline{\tau}_{\text{src}}$. The number shown is the mean of these two macro-averages, written as $\overline{\tau}=\tfrac{1}{2}(\overline{\tau}_{\text{tgt}}+\overline{\tau}_{\text{src}})$. Removing perplexity and image similarity produces the largest drops, while diversity adds complementary coverage.

% \paragraph{Limitations.}
% \name{} is designed for \emph{ranking}; it does not attempt to forecast the absolute \emph{magnitude} of improvement.
% Targets scored with rationale-heavy criteria may require additional features (e.g., rationale similarity) that go beyond QA/visual alignment.

%% file: tables/tao.tex
\begin{table}[t]
  \centering
  \scriptsize
  \setlength{\tabcolsep}{4pt}
  \begin{threeparttable}
    \begin{adjustbox}{max width=\textwidth}
      \begin{tabular}{lcccccccc}
        \toprule
        & \textbf{OCR} & \textbf{Chart} & \textbf{Spatial} & \textbf{Counting}
        & \textbf{General VQA} & \textbf{Document} & \textbf{Math} & \textbf{Avg} \\
        \midrule
        $\tau_{\text{Tgt}}$ &
        \begin{tabular}{@{}l@{}}OCR-VQA: 0.912\\ScreenQA: 0.869\end{tabular} &
        \begin{tabular}{@{}l@{}}ChartQA: 0.911\\Chart2Text: 0.869\end{tabular} &
        \begin{tabular}{@{}l@{}}Open-Spatial: 0.736\\CLEVR(R): 0.824\end{tabular} &
        \begin{tabular}{@{}l@{}}CLEVR(C): 0.846\\TallyQA: 0.822\end{tabular} &
        \begin{tabular}{@{}l@{}}A-OKVQA: 0.868\\VC-GVQA: 0.842\end{tabular} &
        \begin{tabular}{@{}l@{}}DocVQA: 0.890\\FinanceQA: 0.890\end{tabular} &
        \begin{tabular}{@{}l@{}}GeomVerse: 0.911\\MapQA: 0.933\end{tabular} &
        0.863 \\
        \midrule
        $\tau_{\text{Src}}$ &
        \begin{tabular}{@{}l@{}}OCR-VQA: 0.846\\ScreenQA: 0.911\end{tabular} &
        \begin{tabular}{@{}l@{}}ChartQA: 0.869\\Chart2Text: 0.822\end{tabular} &
        \begin{tabular}{@{}l@{}}Open-Spatial: 0.822\\CLEVR(R): 0.846\end{tabular} &
        \begin{tabular}{@{}l@{}}CLEVR(C): 0.869\\TallyQA: 0.846\end{tabular} &
        \begin{tabular}{@{}l@{}}A-OKVQA: 0.846\\VC-GVQA: 0.822\end{tabular} &
        \begin{tabular}{@{}l@{}}DocVQA: 0.869\\FinanceQA: 0.911\end{tabular} &
        \begin{tabular}{@{}l@{}}GeomVerse: 0.846\\MapQA: 0.869\end{tabular} &
        0.857 \\
        \bottomrule
      \end{tabular}
    \end{adjustbox}
    \caption{For each target dataset under the 7 task categories, we compute Kendall’s $\tau$ between the ranking of source datasets by observed relative improvements $\{\Delta_{s\to t}\}_s$ and the ranking induced by the training-free \name{} scores $\{\mathcal{M}(s\!\to\!t)\}_s$. The high correlation coefficients show the effectiveness of \name{} in predicting the contribution ranking of different source datasets.
    % \xingyu{@Xuan change this to two rows, because in evaluation we write two-way tau eval, so here it should have two groups of numbers}
    }
    \label{tab:kendall-by-category}
  \end{threeparttable}
\end{table}
% ------------------------------------------------------

%% file: content/5-data-selection.tex
\section{\name{} for Data Selection}
\label{sec:data-selection}

% Now that we have a training-free metric that could predict a training data's influence on a test data, 
Besides predicting the usefulness ranking of different source datasets, \name{} can also be used
to guide data selection \emph{under a fixed compute budget}. We first introduce two selection
setups for supervised instruction tuning: real-data reweighting (\S\ref{subsec:real-setup}) and
synthetic-data ranking and selection (\S\ref{subsec:synthetic-setup}). We then present the
outcomes and analysis in \S\ref{subsec:selection-results}. Finally, we include an additional
study on RL post-training data selection (\Cref{tab:rl_qwen3b}).
\subsection{Real-data reweighting}
\label{subsec:real-setup}
We fix the total training budget to be $N{=}14\times 20\text{K}=280\text{K}$ samples. Using the fourteen vision–language datasets as described in \Cref{sec:cross_task_influence} as candidate sources, we determine each dataset's relative weighting in the mixture using their \name{} score. Intuitively, datasets assigned higher \name{} scores are expected to contribute more positively to target performance, and are therefore allocated a larger proportion of samples in the mixture. For more details on our reweighting algorithm, please see \Cref{app:detailed-algs}. 
% \lucy{Update this now that we have decided on one D.P. strategy}

% In these scenarios, we use four ways of obtaining the extra data: (1) sample new data from the original full data source (\textbf{D.P. Real}), (2) repeat data randomly from the existing set of 20$K$ (\textbf{D.P. Dup}), and (3) randomly sample data from the synthetic data pool corresponding to this source dataset (\textbf{D.P.-SynthRand}), and (4) select data from synthetic data pool sorted by the instance-level \name{} score (\textbf{D.P.-Synth}). We detail our data synthesis process for curating the synthetic data pool in the section below.
% Weights are normalized and converted to counts with rounding; any residual offset from $N$ is corrected by adjusting the largest buckets by $\pm 1$ until the total matches $N$.

% \paragraph{Implementation Details}

Note that the relative weighting would sometimes require more than 20$K$ items from a given dataset, which is beyond the size of our original 14 datasets. In such scenarios, we sample new data from the original full data source to obtain the desired amount.

For the real-data reweighting setup, we compare the following baseline methods of obtaining the training set of $280\text{K}$ training samples for a given target set.

\textbf{Uniform}: combine the 14 source datasets with $20\text{K}$ examples uniformly sampled from each. 

\textbf{Oracle}: use observed relative improvements from \autoref{fig:rel-heatmap} to reweigh the source datasets. %\lucy{might need to change this name}

\textbf{ICONS}~\citep{wu2025icons}: applies the gradient-based influence consensus approach to score all data in the original real data pool, then use the top $280\text{K}$ datapoints as the training set. This training-required method is current state-of-the-art data selection method for MLLM instruction tuning.

\textbf{D.P.}: use \name{} to reweigh the source datasets.
  % (; extras are additional \emph{real} examples from the corresponding sources.
  % \item \textbf{D.P. Dup} same weights as D.P. Real, but when a source requires more than $20\text{K}$ items, fill the extras by \emph{duplicating} its randomly selected real examples with different sampling seeds rather than drawing new ones.
  % \item \textbf{D.P.-SynthRand} same weights as D.P. Real, but fill any per-source extras with \emph{synthetic} QA pairs sampled \emph{at random} from the synthetic pool built on the same images.
  % \item \textbf{D.P.-Synth} same weights as D.P. Real, but fill any per-source extras with \emph{synthetic} QA pairs selected by the instance-level \name{} score (top-$K$ within the synthetic pool for that source).

\subsection{Synthetic-data selection}
\label{subsec:synthetic-setup} 
Given a target dataset and a large pool of synthetic training candidates, \name{} can also be used to identify the subset of data expected to be most beneficial for improving target performance. We score each individual datapoint $syn\_d$ by a simplified version of the original metric which removes the diversity term:
\begin{equation}
\label{eq:metric-ind}
\mathcal{M}(syn\_d\to\!\mathcal{T})
\;=\;
\frac{\text{QSim}\cdot \text{ASim}\cdot \text{ISim}\cdot \text{PPL}(Syn\_D)}
     {\text{PPL}(\mathcal{T})}.
\end{equation}

In our experiments, we construct the synthetic data pool via the following: For each of the 14 datasets, we split its 20k images into two disjoint halves of 10k images each.
One half is sent to GPT\mbox{-}5 and the other half to Gemini~2.5~Pro. For every image, the generator is asked to produce five VQA-style question–answer pairs conditioned only on the image. Both GPT and Gemini~2.5~Pro receive the \textit{same} prompt template (\Cref{app:prompt-templates}).
This yields $10\text{k}\times 2\times 5 = 100\text{k}$ QA pairs per dataset and $\sim 1.4\text{M}$ pairs in total across the 14 datasets.
We retain all pairs that satisfy basic validity checks (non-empty strings, short answers without special tokens).

Using this synthetic data pool, we compare the following baseline data selection methods: \textbf{Uniform}, which uniformly samples 280k data from the full synthetic data pool and \textbf{ICONS}~\citep{wu2025icons}, to our method \textbf{D.P.} for \name.

% \paragraph{Training and selection alignment}
% Training recipe and selection strategies mirror the real-data setup in \Cref{subsec:real-setup}: we fine-tune InternVL3\mbox{-}2B for one epoch using full-parameter SFT implemented in LLaMA\mbox{-}Factory~\citep{zheng2024llamafactory}, and we keep all hyperparameters identical.
% Where a strategy requires \textit{extra} samples beyond the 20k real anchor, the extras for a given source are drawn from that source’s synthetic pool either at random or via the instance-level \name{} scoring, exactly matching the corresponding real-data strategy semantics.

\input{tables/data_selection_results}

\input{tables/rl_qwen3b_grpo}

\subsection{Results and Analysis}
\label{subsec:selection-results}
\paragraph{\name{} provides effective training-free guidance for data selection.}
For each of the 14 target benchmark datasets, we conduct real-data reweighting and synthetic-data
selection using the methods above. \Cref{tab:data_selection_results} summarizes the results for
both selection paradigms. Overall, \name{}-guided selection achieves the best macro-average
performance across tasks in both the real and synthetic setups, yielding 3.4\% and 6.9\%
improvements, respectively. Notably, reweighting guided by \name{} matches or exceeds the
\textbf{Oracle} reweighting constructed from observed performance gains. Moreover, our
training-free metric outperforms the training-based selector ICONS, demonstrating that \name{}
is both effective and computationally efficient for data selection.

\paragraph{Does Gemini 2.5 Pro or GPT 5 provide higher-quality supervision data?} Interestingly, it turns out that Gemini produces higher-quality synthetic data than GPT.
In the synthetic data setup, about $38\%$ of selected items come from GPT\mbox{-}5 and $62\%$ from Gemini~2.5~Pro, indicating that Gemini contributes a larger share of items that score highly under the instance-level \name{} metric. Manual inspection suggests that these items are more challenging, reflected by higher per-token perplexity, and better aligned with the target’s distributions in layout, entities, and answer forms.

\paragraph{\name{} also improves RL post-training data selection.}
We further evaluate whether \name{} can guide data selection for RL post-training.
We perform GRPO training using the \texttt{verl} implementation on Qwen2.5VL-3B-Instruct with
default hyperparameters, and fix the RL prompt budget to $300{\times}14$.
We compare equal allocation across the 14 datasets to a \name{}-guided allocation for real RL
data, and random sampling to \name{}-guided selection for synthetic RL data.
As shown in \Cref{tab:rl_qwen3b}, \name{} improves the average score from 0.583 to 0.595 for
real RL data and from 0.564 to 0.577 for synthetic RL data on the 14-benchmark test suite.

%% file: tables/data_selection_results.tex
\begin{table*}[t!]
    \centering
    \begin{adjustbox}{max width=\textwidth}
    \setlength{\tabcolsep}{5pt}
    \fontsize{9pt}{13.5pt}\selectfont
    \begin{tabular}{l|ccccccc}
        \toprule[1.2pt]
        & \multicolumn{4}{c}{\textbf{Real Data Reweighting}} 
        & \multicolumn{3}{c}{\textbf{Synthetic Data Selection}} \\
        \cmidrule(lr){2-5} \cmidrule(lr){6-8} 
        \multirow{2}{*}{\textbf{Benchmarks / Sampling Methods}} & \multicolumn{2}{c}{\textit{Baseline}} & \multicolumn{1}{c}{\textit{Gold}} 
         & \textit{Ours} 
         & \multicolumn{2}{c}{\textit{Baseline}} 
         & \textit{Ours} \\
        \cmidrule(lr){2-3} \cmidrule(lr){4-4} \cmidrule(lr){5-5} \cmidrule(lr){6-7} \cmidrule(lr){8-8}
         & Uniform & ICONS & Oracle & \textsc{D.P.} & Uniform & ICONS & \textsc{D.P.} \\
        \midrule[1.2pt]
        OCR-VQA~\citep{mishra2019ocr}     & 85.2 & 86.5 & 87.5 & 87.4 & 65.4 & 73.5 & 74.6 \\
        ScreenQA~\citep{hsiao2022screenqa}    & 83.1 & 83.6 & 84.8 & 85.0 & 67.3 & 76.5 & 78.4 \\
        ChartQA~\citep{masry2022chartqa}     & 83.7 & 84.8 & 84.9 & 85.6 & 68.2 & 77.2 & 78.3 \\
        Chart2Text~\citep{kantharaj2022chart}  & 84.0 & 84.5 & 86.0 & 86.2 & 68.7 & 75.9 & 77.5 \\
        Open-Spatial~\citep{cheng2024spatialrgpt}& 37.9 & 40.7 & 41.2 & 41.1 & 26.7 & 27.3 & 28.5 \\
        CLEVR(R)~\citep{johnson2017clevr}    & 41.5 & 42.2 & 45.0 & 45.6 & 33.8 & 35.2 & 37.6 \\
        CLEVR(C)~\citep{johnson2017clevr}    & 44.5 & 45.3 & 47.2 & 48.1 & 35.6 & 36.8 & 38.6 \\
        TallyQA~\citep{acharya2019tallyqa}     & 41.8 & 42.8 & 45.2 & 45.1 & 30.4 & 35.5 & 36.2 \\
        A-OKVQA~\citep{schwenk2022aokvqa}     & 83.7 & 84.9 & 85.8 & 86.0 & 68.2 & 74.4 & 74.8 \\
        VC-GVQA~\citep{shao2024visualcot}     & 81.5 & 82.3 & 83.7 & 83.5 & 67.8 & 75.9 & 76.8 \\
        DocVQA~\citep{Mathew_2021_WACV}      & 76.9 & 79.4 & 80.4 & 81.4 & 58.6 & 66.4 & 67.2 \\
        FinanceQA~\citep{sujetfinanceqa2024}   & 73.8 & 74.1 & 75.1 & 75.2 & 57.2 & 64.8 & 65.9 \\
        GeomVerse~\citep{kazemi2023geomverse}   & 60.1 & 61.2 & 62.4 & 63.0 & 47.1 & 53.1 & 54.4 \\
        MapQA~\citep{chang2022mapqa}       & 82.9 & 83.3 & 83.5 & 84.8 & 75.2 & 78.2 & 79.8 \\
        \midrule
        \textit{Average}        & 67.6 & 69.6 & 70.8 & \textbf{71.0} & 55.1 &  60.8 &  \textbf{62.0} \\
        \textit{Improve} over Uniform & - & +2.0 & +3.2 & \textbf{+3.4} & - & +5.7 & \textbf{+6.9}\\
        \bottomrule[1.2pt]
    \end{tabular}%
    \end{adjustbox}
    \caption{\textbf{Data selection results.} Keeping a fixed compute budget of 280K datapoints, we present the data selection results on 14 target benchmarks. Sampling methods: \texttt{Uniform} is the uniform sampling baseline, \texttt{ICONS} is the sota training-based baseline, and \texttt{Oracle} is the gold baseline that selects samples based on oracle real experiment performance improvement as in \Cref{fig:rel-heatmap}. \texttt{Average} reports the mean performance across all benchmarks, and \textit{Improve over Uniform} shows the absolute performance gain relative to uniform sampling. 
    From the results, our \name{}-based sampling achieves the best average performance on both real and synthetic data, surprisingly reaching up to +0.2\% improvement over the gold oracle selection.}
    \label{tab:data_selection_results}
\end{table*}

%% file: tables/rl_qwen3b_grpo.tex
% RL table styled to match Table~\ref{tab:data_selection_results}
\begin{table*}[t!]
    \centering
    \begin{adjustbox}{max width=\textwidth}
    \setlength{\tabcolsep}{6pt}
    \fontsize{9pt}{13.5pt}\selectfont
    \begin{tabular}{l|c|cc|cc}
        \toprule[1.2pt]
        & \multicolumn{1}{c|}{\textbf{No RL}}
        & \multicolumn{2}{c|}{\textbf{Real RL Data Selection}}
        & \multicolumn{2}{c}{\textbf{Synthetic RL Data Selection}} \\
        \cmidrule(lr){2-2} \cmidrule(lr){3-4} \cmidrule(lr){5-6}
        \multirow{2}{*}{\textbf{Benchmarks / Methods}}
        & \textit{Baseline}
        & \multicolumn{1}{c}{\textit{Baseline}} & \textit{Ours}
        & \multicolumn{1}{c}{\textit{Baseline}} & \textit{Ours} \\
        \cmidrule(lr){2-2} \cmidrule(lr){3-3} \cmidrule(lr){4-4} \cmidrule(lr){5-5} \cmidrule(lr){6-6}
        & Baseline & Equal & \textsc{D.P.} & Random & \textsc{D.P.} \\
        \midrule[1.2pt]
        OCR-VQA~\citep{mishra2019ocr}            & 72 & 75 & 76 & 71 & 73 \\
        ScreenQA~\citep{hsiao2022screenqa}       & 66 & 68 & 69 & 68 & 67 \\
        ChartQA~\citep{masry2022chartqa}         & 70 & 71 & 73 & 69 & 72 \\
        Chart2Text~\citep{kantharaj2022chart}    & 68 & 70 & 72 & 70 & 69 \\
        Open-Spatial~\citep{cheng2024spatialrgpt}& 35 & 39 & 41 & 36 & 37 \\
        CLEVR(R)~\citep{johnson2017clevr}        & 41 & 44 & 46 & 42 & 42 \\
        CLEVR(C)~\citep{johnson2017clevr}        & 49 & 52 & 52 & 50 & 52 \\
        TallyQA~\citep{acharya2019tallyqa}       & 31 & 34 & 33 & 33 & 35 \\
        A-OKVQA~\citep{schwenk2022aokvqa}        & 64 & 68 & 69 & 62 & 65 \\
        VC-GVQA~\citep{shao2024visualcot}        & 65 & 67 & 69 & 64 & 65 \\
        DocVQA~\citep{Mathew_2021_WACV}          & 59 & 62 & 60 & 62 & 63 \\
        FinanceQA~\citep{sujetfinanceqa2024}     & 54 & 55 & 56 & 55 & 54 \\
        GeomVerse~\citep{kazemi2023geomverse}    & 42 & 44 & 47 & 43 & 45 \\
        MapQA~\citep{chang2022mapqa}             & 64 & 67 & 66 & 64 & 68 \\
        \midrule
        \textit{Average} & 55.7 & 58.3 & \textbf{59.5} & 56.4 & \textbf{57.7} \\
        \textit{Improve} over Baseline
        & - & +2.6 & \textbf{+3.8} & +0.7 & \textbf{+2.0} \\
        \bottomrule[1.2pt]
    \end{tabular}%
    \end{adjustbox}
    \caption{\textbf{RL post-training data selection results.} We fix a prompt budget of $300{\times}14$ and evaluate on the same 14-benchmark test suite. For real RL data, we compare equal allocation (\text{Equal}) to \name{}-guided allocation (\textsc{D.P.}). For synthetic RL data, we compare random sampling (\text{Random}) to \name{}-guided selection. \textit{Average} reports the mean score across benchmarks, and \textit{Improve over Baseline} shows the absolute gain relative to the no-RL baseline.}
    \label{tab:rl_qwen3b}
\end{table*}

%% file: content/2-related.tex
\section{Related Works}
% \xingyu{Include: DataComp: In search of the next generation of multimodal datasets, When less is more: Investigating data pruning for pretraining llms at scale, IMPROVING PRETRAINING DATA USING PERPLEXITY CORRELATIONS.
% }
\paragraph{Data Mixtures for LMs}
Data Mixture has long been an interesting problem in the research community. 
% https://arxiv.org/pdf/2305.10429
% DoReMi uses a proxy model...during pretraining...\\ 
\citet{xie2023doremi} propose DoReMi, which trains a small proxy model with GroupDRO to learn domain weights and then resamples the corpus for full-scale training.
%LESS got rid of the proxy model, and deploys an Adam gradient similarity approach... during instruction finetuning stage... \\
A complementary line fits predictive mixture laws~\citep{ye2025dml, ge2025bimix, kang2025autoscale, li2025dmo}. Beyond domain-level mixtures, example-level selection has been studied for both pretraining and instruction tuning, such as DSIR~\citep{xie2023dsir} and LESS~\citep{xia2024less}. Simple perplexity-based pruning and perplexity correlations offer strong, training-free selectors at pretraining scale~\citep{marion2023less,thrush2024perplexity, qi2025difficulty}. Our work differs in being training-free: instead of proxy runs, gradients, or online adaptation, we propose a simple metric that scores training sets for multi-task utility before any finetuning.
% https://openreview.net/pdf?id=jjCB27TMK3
%Data Mixing Law .... pretraining...

\paragraph{Data Mixtures for Multimodal LMs}
% from Icons:
% Prior work has explored various data selection strategies, including gradient-based approaches [45,6], influence functions [47, 19], and diversity-based sampling [48, 4]. However, many of these methods either optimize for single tasks in isolation or maximize source diversity without aligning to downstream needs. In multitask visual instruction tuning, this is particularly limiting: optimizing for one task may hurt generalization, and task-agnostic diversity may dilute impact. Rather than selecting data based on per-task influence, we aim to identify samples that are broadly useful—training examples that consistently contribute across tasks. To do this, we aggregate gradient-based influence scores using a simple yet effective majority voting scheme.
% https://arxiv.org/pdf/2501.00654
% ICONS also gradient based. But focused on selecting less data to achieve approximate performance. \\
In instruction tuning of VLMs, ICONS aggregates first-order influence estimates across tasks to identify broadly useful examples, attaining near-parity with full-data training using compact subsets~\citep{wu2025icons}. Related selection frameworks estimate task- and instance-level value using gradient features~\citep{liu2024tive} or score with the model itself to filter hard/diverse instructions~\citep{chen2024selffilter}. On the curation side, \textsc{DataComp} standardizes large-scale image–text pretraining as a dataset filtering problem, enabling fair comparison across selection strategies and scales~\citep{gadre2023datacomp}.
% https://arxiv.org/pdf/2505.24871
At the post-training/RL stage, MoDoMoDo formulates multimodal RLVR as a mixture-optimization problem, learning policies that improve out-of-distribution generalization~\citep{liang2025modomodo}. In contrast to these training-dependent paradigms (proxy sweeps, gradient stores, or RL rollouts), we target a training-free, pre-finetuning decision rule that is compute-light and directly applicable to both real and synthetic multimodal data.
% MoDoMoDo estimating the optimal data mixture strategy with training a non-linear second-order model during RL stage.

% \xingyu{add hangfeng's paper}

\paragraph{Domain and dataset generalization for LMs} In the LM context, existing work explores the generalizability of SFT or RL training from one domain to another. Reinforcement post-training (RPT) can generalize beyond the fine-tuned domain under certain conditions~\citep{hu2025rpt}; comparative studies suggest SFT tends to memorize while RL-style post-training generalizes more robustly~\citep{chu2025sftvsrl}; and math-specialized training shows mixed transfer unless optimized via RL rather than SFT~\citep{huan2025mathgeneralization}. However, these works do not offer metrics for analyzing the contribution of different datasets to improving performance on the target domain. Our work seeks to both deepen understanding and offer prescriptive guidance. 
%https://arxiv.org/pdf/2506.19733
%https://arxiv.org/abs/2507.00432

%% file: content/6-conclusion.tex
\section{Conclusions}
% \xingyu{@Xuan. add one by throwing abstract and intro to gpt.}

In this paper, we investigate whether one can predict before any training how much a given source dataset will benefit a target dataset in MLLMs. Across seven task families, we find that intuitive task similarity is an unreliable guide and that transfer is dataset-specific rather than task-specific. To address this, we introduce \name{}, a simple, training-free and interpretable metric that integrates cross-modal similarity, multimodal perplexity, and source dataset diversity. \name{} closely aligns with realized training outcomes when ranking data influence, and further serves as a useful guidance for both real and synthetic data selection. We believe \name{} provides a principled signal for data curation, enabling more effective MLLM training.

%% file: content/appendix.tex
\section{Appendix}

\subsection{The Use of LLMs}

LLMs did not play an important role in this paper’s research ideation or writing to the extent that they should be regarded as a contributor. In the experiments, LLMs are the main experimental object.

\subsection{Implementation Details for \autoref{sec:cross_task_influence}}
\label{appen:implementation-details}

All runs fine-tune the base model for one epoch using full-parameter supervised fine-tuning (Full SFT) implemented in LLaMA-Factory~\citep{zheng2024llamafactory}. The learning rate is set to be $1e-5$. All evaluations follow the benchmarks' task-appropriate metrics.

\subsection{Detailed Algorithms}
\label{app:detailed-algs}
\begin{algorithm}[h]
\caption{Data Reweighting based on the metric $\mathcal{M}$}
\label{alg:data-reweight}
\begin{algorithmic}[1]
\Require target dataset $\mathcal{D}_t$; source datasets $\{\mathcal{D}_s^{(i)}\}_{i=1}^{M}$; total budget $N$
\Ensure training set $S$ for SFT
\For{$i \gets 1$ \textbf{to} $M$}
    \State $m_i \gets \mathcal{M}\!\big(\mathcal{D}_s^{(i)} \!\to\! \mathcal{D}_t\big)$  \Comment{compute metric between source $i$ and target}
\EndFor
\State $Z \gets \sum_{j=1}^{M} m_j$
\For{$i \gets 1$ \textbf{to} $M$}
    \State $N_i \gets \operatorname{round}\!\Big(N \cdot \frac{m_i}{Z}\Big)$
    \State $S_i \gets \textsc{SampleUniform}\big(\mathcal{D}_s^{(i)},\, N_i\big)$
\EndFor
\State $S \gets \bigcup_{i=1}^{M} S_i$
\State \Return $S$
\end{algorithmic}
\end{algorithm}

\begin{algorithm}[h]
\caption{Metric-guided Synthetic Data Selection}
\label{alg:synthetic-selection}
\begin{algorithmic}[1]
\Require target dataset $\mathcal{D}_t$; synthetic pool $\mathcal{S}$; either Top-$K$ or threshold $\tau$
\Ensure selected synthetic subset $\mathcal{S}_{\text{sel}}$
\For{\textbf{each} $u \in \mathcal{S}$}
    \State $\text{QSim}(u,\mathcal{D}_t),\, \text{ASim}(u,\mathcal{D}_t),\, \text{ISim}(u,\mathcal{D}_t)$ 
    \State $\text{PPL}_{\text{train}}(\{u\}),\, \text{PPL}_{\text{test}}(\mathcal{D}_t)$ 
    \State $\tilde{m}(u) \gets \text{QSim}\!\cdot\!\text{ASim}\!\cdot\!\text{ISim}\!\cdot\!\text{PPL}_{\text{train}}(\{u\})\,/\,\text{PPL}_{\text{test}}(\mathcal{D}_t)$ 
\EndFor
\If{\textsc{UseTopK}}
    \State $\mathcal{S}_{\text{sel}} \gets \textsc{TopK\_By\_Value}\big(\mathcal{S},\, \tilde{m},\, K\big)$ 
\Else
    \State $\mathcal{S}_{\text{sel}} \gets \{\, u \in \mathcal{S} : \tilde{m}(u) \ge \tau \,\}$ 
\EndIf
\State \Return $\mathcal{S}_{\text{sel}}$
\end{algorithmic}
\end{algorithm}

\subsection{Prompt Templates}
\label{app:prompt-templates}
\noindent The following template is used for both GPT\mbox{-}5 and Gemini~2.5~Pro.
The calling code injects the image in place of \texttt{<IMAGE>}. 
% \lucy{move detailed prompts to appendix.}

\begin{quote}\small\ttfamily
You are generating visual question answering data strictly grounded in the given image. Use only what is visible in the image. Do not rely on outside knowledge. These questions should be inference questions about what is in the picture. \\[0.25em]
Image: \textless IMAGE\textgreater\\
Task: Create exactly five diverse VQA pairs about this image.\\
    Constraints: questions must be answerable solely from the image; answers should be short (several words or a number); avoid ambiguous or subjective wording; avoid near-duplicate questions; avoid requiring reading tiny illegible text; if no text is legible, do not ask OCR questions.\\
Output format: return a single JSON array of five objects. Each object has the following fields\\
\hspace*{1.5em}\{-\} \texttt{"question"}: one string\\
\hspace*{1.5em}\{-\} \texttt{"answer"}: one string (lowercase for each word and numerals for numbers)\\
Do not include any additional text before or after the JSON.\\[0.25em]
\end{quote}

%% file: iclr2026_conference.bib
@String(ICLR = {Int. Conf. Learn. Represent.})

@String(AAAI = {AAAI})

@String(ICLR  = {ICLR})

@article{thrush2024improving,
  title={Improving pretraining data using perplexity correlations},
  author={Thrush, Tristan and Potts, Christopher and Hashimoto, Tatsunori},
  journal={arXiv preprint arXiv:2409.05816},
  year={2024}
}

@article{marion2023less,
  title={When less is more: Investigating data pruning for pretraining llms at scale},
  author={Marion, Max and {\"U}st{\"u}n, Ahmet and Pozzobon, Luiza and Wang, Alex and Fadaee, Marzieh and Hooker, Sara},
  journal={arXiv preprint arXiv:2309.04564},
  year={2023}
}

@article{li2023quantity,
  title={From quantity to quality: Boosting llm performance with self-guided data selection for instruction tuning},
  author={Li, Ming and Zhang, Yong and Li, Zhitao and Chen, Jiuhai and Chen, Lichang and Cheng, Ning and Wang, Jianzong and Zhou, Tianyi and Xiao, Jing},
  journal={arXiv preprint arXiv:2308.12032},
  year={2023}
}

@article{albalak2024survey,
  title={A survey on data selection for language models},
  author={Albalak, Alon and Elazar, Yanai and Xie, Sang Michael and Longpre, Shayne and Lambert, Nathan and Wang, Xinyi and Muennighoff, Niklas and Hou, Bairu and Pan, Liangming and Jeong, Haewon and others},
  journal={arXiv preprint arXiv:2402.16827},
  year={2024}
}

@article{sachdeva2024train,
  title={How to train data-efficient llms},
  author={Sachdeva, Noveen and Coleman, Benjamin and Kang, Wang-Cheng and Ni, Jianmo and Hong, Lichan and Chi, Ed H and Caverlee, James and McAuley, Julian and Cheng, Derek Zhiyuan},
  journal={arXiv preprint arXiv:2402.09668},
  year={2024}
}

@article{cheng2024spatialrgpt,
  title={Spatialrgpt: Grounded spatial reasoning in vision-language models},
  author={Cheng, An-Chieh and Yin, Hongxu and Fu, Yang and Guo, Qiushan and Yang, Ruihan and Kautz, Jan and Wang, Xiaolong and Liu, Sifei},
  journal={Advances in Neural Information Processing Systems},
  volume={37},
  pages={135062--135093},
  year={2024}
}

@article{bai2025qwen2.5vl,
  title={Qwen2. 5-vl technical report},
  author={Bai, Shuai and Chen, Keqin and Liu, Xuejing and Wang, Jialin and Ge, Wenbin and Song, Sibo and Dang, Kai and Wang, Peng and Wang, Shijie and Tang, Jun and others},
  journal={arXiv preprint arXiv:2502.13923},
  year={2025}
}

@article{xia2024less,
  title={Less: Selecting influential data for targeted instruction tuning},
  author={Xia, Mengzhou and Malladi, Sadhika and Gururangan, Suchin and Arora, Sanjeev and Chen, Danqi},
  journal={arXiv preprint arXiv:2402.04333},
  year={2024}
}

@article{team2023gemini,
  title={Gemini: a family of highly capable multimodal models},
  author={Team, Gemini and Anil, Rohan and Borgeaud, Sebastian and Wu, Yonghui and Alayrac, Jean-Baptiste and Yu, Jiahui and Soricut, Radu and Schalkwyk, Johan and Dai, Andrew M and Hauth, Anja and others},
  journal={arXiv preprint arXiv:2312.11805},
  year={2023}
}

@article{zheng2024llamafactory,
  title   = {LlamaFactory: Unified Efficient Fine-Tuning of 100+ Language Models},
  author  = {Zheng, Yaowei and Zhang, Richong and Zhang, Junhao and Ye, Yanhan and Luo, Zheyan and Feng, Zhangchi and Ma, Yongqiang},
  journal = {arXiv preprint arXiv:2403.13372},
  year    = {2024},
  note    = {ACL 2024 System Demonstration Track},
  doi     = {10.48550/arXiv.2403.13372},
  url     = {https://arxiv.org/abs/2403.13372}
}

@article{thrush2024perplexity,
  title   = {Improving Pretraining Data Using Perplexity Correlations},
  author  = {Thrush, Tristan and Potts, Christopher and Hashimoto, Tatsunori},
  journal = {arXiv preprint arXiv:2409.05816},
  year    = {2024},
  note    = {ICLR 2025},
  url     = {https://arxiv.org/abs/2409.05816}
}

@article{qi2025difficulty,
  title={Difficulty-Based Preference Data Selection by DPO Implicit Reward Gap},
  author={Qi, Xuan and Xu, Rongwu and Jin, Zhijing},
  journal={arXiv preprint arXiv:2508.04149},
  year={2025}
}

@article{wang2024diversitySubsetSelection,
  title   = {Diversity Measurement and Subset Selection for Instruction Tuning Datasets},
  author  = {Wang, Peiqi and Shen, Yikang and Guo, Zhen and Stallone, Matthew and Kim, Yoon and Golland, Polina and Panda, Rameswar},
  journal = {arXiv preprint arXiv:2402.02318},
  year    = {2024},
  url     = {https://arxiv.org/abs/2402.02318}
}

@misc{gpt4,
    title={GPT-4 Technical Report},
    author={OpenAI},
    year={2023},
    eprint={2303.08774},
    archivePrefix={arXiv},
    primaryClass={cs.CL}
}

@misc{bai2023qwenvl,
    title={Qwen-VL: A Versatile Vision-Language Model for Understanding, Localization, Text Reading, and Beyond},
    author={Jinze Bai and Shuai Bai and Shusheng Yang and Shijie Wang and Sinan Tan and Peng Wang and Junyang Lin and Chang Zhou and Jingren Zhou},
    year={2023},
    eprint={2308.12966},
    archivePrefix={arXiv},
    primaryClass={cs.CV}
}

@article{qwen2vl2024,
  title={Qwen2-VL: Enhancing Vision-Language Model's Perception of the World at Any Resolution},
  author={Wang, Peng and Bai, Shuai and Tan, Sinan and Wang, Shijie and Fan, Zhihao and Bai, Jinze and Chen, Keqin and Liu, Xuejing and Wang, Jialin and Ge, Wenbin and others},
  journal={arXiv preprint arXiv:2409.12191},
  year={2024}
}

@inproceedings{mishra2019ocr,
  title={OCR-VQA: Visual Question Answering by Reading Text in Images},
  author={Mishra, Anand and Shekhar, Shashank and Singh, Ajeet Kumar and Chakraborty, Anirban},
  booktitle={Proceedings of the IEEE/CVF International Conference on Computer Vision},
  pages={4492--4501},
  year={2019}
}

@inproceedings{hsiao2022screenqa,
  title={ScreenQA: Large-Scale Question-Answer Pairs over Mobile App Screenshots},
  author={Hsiao, Yao-Hung Hubert and Wu, Yue and Chen, Xiaojuan and Lee, Tak-Sang},
  booktitle={Proceedings of the 2022 Conference on Empirical Methods in Natural Language Processing},
  pages={8384--8395},
  year={2022}
}

@inproceedings{masry2022chartqa,
  title={ChartQA: A Benchmark for Question Answering about Charts with Visual and Logical Reasoning},
  author={Masry, Ahmed and Do, Xuan Long and Joty, Shafiq and Hoque, Enamul},
  booktitle={Findings of the Association for Computational Linguistics: ACL 2022},
  pages={2263--2279},
  year={2022}
}

@inproceedings{kantharaj2022chart,
  title={Chart-to-text: A large-scale benchmark for chart summarization},
  author={Kantharaj, Shankar and Leong, Rixie Tian Qi and Lin, Xiang and Masry, Ahmed and Thakkar, Megh and Hoque, Enamul and Joty, Shafiq},
  booktitle={Proceedings of the 60th Annual Meeting of the Association for Computational Linguistics (Volume 1: Long Papers)},
  pages={4005--4023},
  year={2022}
}

@inproceedings{johnson2017clevr,
  title={CLEVR: A Diagnostic Dataset for Compositional Language and Elementary Visual Reasoning},
  author={Johnson, Justin and Hariharan, Bharath and van der Maaten, Laurens and Fei-Fei, Li and Lawrence Zitnick, C and Girshick, Ross},
  booktitle={Proceedings of the IEEE Conference on Computer Vision and Pattern Recognition},
  pages={2901--2910},
  year={2017}
}

@inproceedings{acharya2019tallyqa,
  title={TallyQA: Answering Complex Counting Questions},
  author={Acharya, Manoj and Kafle, Kushal and Kanan, Christopher},
  booktitle={Proceedings of the AAAI Conference on Artificial Intelligence},
  volume={33},
  pages={8076--8084},
  year={2019}
}

@article{rousseeuw1987silhouettes,
  title        = {Silhouettes: A graphical aid to the interpretation and validation of cluster analysis},
  author       = {Rousseeuw, Peter J.},
  journal      = {Journal of Computational and Applied Mathematics},
  volume       = {20},
  number       = {1--2},
  pages        = {53--65},
  year         = {1987},
  doi          = {10.1016/0377-0427(87)90125-7}
}

@article{schwenk2022aokvqa,
  title   = {A-OKVQA: A Benchmark for Visual Question Answering using World Knowledge},
  author  = {Schwenk, Dustin and Khandelwal, Apoorv and Clark, Christopher and Marino, Kenneth and Mottaghi, Roozbeh},
  journal = {arXiv preprint arXiv:2206.01718},
  year    = {2022}
}

@article{shao2024visualcot,
  title   = {Visual CoT: Advancing Multi-Modal Language Models with a Comprehensive Dataset and Benchmark for Chain-of-Thought Reasoning},
  author  = {Shao, Hao and Qian, Shengju and Xiao, Han and Song, Guanglu and Zong, Zhuofan and Wang, Letian and Liu, Yu and Li, Hongsheng},
  journal = {arXiv preprint arXiv:2403.16999},
  year    = {2024},
  doi     = {10.48550/arXiv.2403.16999}
}

@inproceedings{Mathew_2021_WACV,
  author    = {Mathew, Minesh and Karatzas, Dimosthenis and Jawahar, C.~V.},
  title     = {DocVQA: A Dataset for VQA on Document Images},
  booktitle = {Proceedings of the IEEE/CVF Winter Conference on Applications of Computer Vision (WACV)},
  year      = {2021},
  month     = {January},
  pages     = {2200--2209}
}

@misc{sujetfinanceqa2024,
  title        = {Sujet-Finance-QA-Vision-100k},
  author       = {{Sujet AI}},
  year         = {2024},
  howpublished = {Hugging Face Dataset},
  url          = {https://huggingface.co/datasets/sujet-ai/Sujet-Finance-QA-Vision-100k},
  note         = {Accessed: 2025-09-03}
}

@article{kazemi2023geomverse,
  title   = {GeomVerse: A Systematic Evaluation of Large Models for Geometric Reasoning},
  author  = {Kazemi, Mehran and Alvari, Hamidreza and Anand, Ankit and Wu, Jialin and Chen, Xi and Soricut, Radu},
  journal = {arXiv preprint arXiv:2312.12241},
  year    = {2023}
}

@article{chang2022mapqa,
  title   = {MapQA: A Dataset for Question Answering on Choropleth Maps},
  author  = {Chang, Shuaichen and Palzer, David and Li, Jialin and Fosler-Lussier, Eric and Xiao, Ningchuan},
  journal = {arXiv preprint arXiv:2211.08545},
  year    = {2022}
}

@inproceedings{xie2023doremi,
  title     = {DoReMi: Optimizing Data Mixtures Speeds Up Language Model Pretraining},
  author    = {Xie, Sang Michael and Pham, Hieu and Dong, Xuanyi and Du, Nan and Liu, Hanxiao and Lu, Yifeng and Liang, Percy and Le, Quoc V. and Ma, Tengyu and Yu, Adams Wei},
  booktitle = {Advances in Neural Information Processing Systems (NeurIPS)},
  year      = {2023},
  url       = {https://arxiv.org/abs/2305.10429}
}

@inproceedings{ye2025dml,
  title     = {Data Mixing Laws: Optimizing Data Mixtures by Predicting Language Modeling Performance},
  author    = {Ye, Jiasheng and Liu, Peiju and Sun, Tianxiang and Zhan, Jun and Zhou, Yunhua and Qiu, Xipeng},
  booktitle = {International Conference on Learning Representations (ICLR)},
  year      = {2025},
  url       = {https://openreview.net/forum?id=jjCB27TMK3}
}

@misc{ge2025bimix,
  title        = {BiMix: Bivariate Data Mixing Law for Language Model Pretraining},
  author       = {Ge, Ce and Ma, Zhijian and Chen, Daoyuan and Li, Yaliang and Ding, Bolin},
  year         = {2025},
  howpublished = {OpenReview preprint},
  note         = {Submitted to ICLR 2025},
  url          = {https://openreview.net/forum?id=JsM46OZix7}
}

@inproceedings{kang2025autoscale,
  title     = {AutoScale: Scale-Aware Data Mixing for Pre-Training LLMs},
  author    = {Kang, Feiyang and Sun, Yifan and Wen, Bingbing and Chen, Si and Song, Dawn and Mahmood, Rafid and Jia, Ruoxi},
  booktitle = {Conference on Language Modeling (COLM)},
  year      = {2025},
  url       = {https://openreview.net/forum?id=rujwIvjooA}
}

@inproceedings{xie2023dsir,
  title     = {Data Selection for Language Models via Importance Resampling},
  author    = {Xie, Sang Michael and Santurkar, Shibani and Ma, Tengyu and Liang, Percy},
  booktitle = {Advances in Neural Information Processing Systems (NeurIPS)},
  year      = {2023},
  url       = {https://arxiv.org/abs/2302.03169}
}

@article{wu2025icons,
  title   = {{ICONS}: Influence Consensus for Vision-Language Data Selection},
  author  = {Wu, Xindi and Xia, Mengzhou and Shao, Rulin and Deng, Zhiwei and Koh, Pang Wei and Russakovsky, Olga},
  journal = {arXiv preprint arXiv:2501.00654},
  year    = {2025},
  url     = {https://arxiv.org/abs/2501.00654}
}

@article{liu2024tive,
  title   = {Less is More: High-value Data Selection for Visual Instruction Tuning},
  author  = {Liu, Zikang and Zhou, Kun and Zhao, Wayne Xin and Gao, Dawei and Li, Yaliang and Wen, Ji-Rong},
  journal = {arXiv preprint arXiv:2403.09559},
  year    = {2024},
  url     = {https://arxiv.org/abs/2403.09559}
}

@article{chen2024selffilter,
  title   = {Your Vision-Language Model Itself Is a Strong Filter: Towards High-Quality Instruction Tuning with Data Selection},
  author  = {Chen, Ruibo and Wu, Yihan and Chen, Lichang and Liu, Guodong and He, Qi and Xiong, Tianyi and Liu, Chenxi and Guo, Junfeng and Huang, Heng},
  journal = {arXiv preprint arXiv:2402.12501},
  year    = {2024},
  url     = {https://arxiv.org/abs/2402.12501}
}

@inproceedings{gadre2023datacomp,
  title     = {DataComp: In Search of the Next Generation of Multimodal Datasets},
  author    = {Gadre, Samir Yitzhak and Ilharco, Gabriel and Fang, Alex and Hayase, Jonathan and Smyrnis, Georgios and Nguyen, Thao and Marten, Ryan and Wortsman, Mitchell and Ghosh, Dhruba and Zhang, Jieyu and Orgad, Eyal and Entezari, Rahim and Daras, Giannis and Pratt, Sarah and Ramanujan, Vivek and Bitton, Yonatan and Marathe, Kalyani and Mussmann, Stephen and Vencu, Richard and Cherti, Mehdi and Krishna, Ranjay and Saukh, Olga and Ratner, Alexander and Song, Shuran and Hajishirzi, Hannaneh and Farhadi, Ali and Beaumont, Romain and Dimakis, Alexandros and Jitsev, Jenia and Carmon, Yair and Shankar, Vaishaal and Schmidt, Ludwig},
  booktitle = {Advances in Neural Information Processing Systems (NeurIPS)},
  year      = {2023},
  url       = {https://arxiv.org/abs/2304.14108}
}

@article{liang2025modomodo,
  title   = {MoDoMoDo: Multi-Domain Data Mixtures for Multimodal LLM Reinforcement Learning},
  author  = {Liang, Yiqing and Qiu, Jielin and Ding, Wenhao and Liu, Zuxin and Tompkin, James and Xu, Mengdi and Xia, Mengzhou and Tu, Zhengzhong and Shi, Laixi and Zhu, Jiacheng},
  journal = {arXiv preprint arXiv:2505.24871},
  year    = {2025},
  url     = {https://arxiv.org/abs/2505.24871}
}

@article{chu2025sftvsrl,
  title   = {SFT Memorizes, RL Generalizes: A Comparative Study of Foundation Model Post-training},
  author  = {Chu, Tianzhe and Zhai, Yuexiang and Yang, Jihan and Tong, Shengbang and Xie, Saining and Schuurmans, Dale and Le, Quoc V. and Levine, Sergey and Ma, Yi},
  journal = {arXiv preprint arXiv:2501.17161},
  year    = {2025},
  url     = {https://arxiv.org/abs/2501.17161}
}

@article{huan2025mathgeneralization,
  title   = {Does Math Reasoning Improve General LLM Capabilities? A Controlled Study of Transfer and Forgetting},
  author  = {Huan, Mingzhen and others},
  journal = {arXiv preprint arXiv:2507.00432},
  year    = {2025},
  url     = {https://arxiv.org/abs/2507.00432}
}

@article{hu2025rpt,
  title   = {Breaking Barriers: Do Reinforcement Post Training Gains Transfer To Unseen Domains?},
  author  = {Hu, Chuxuan and Zhu, Yuxuan and Kellermann, Antony and Biddulph, Caleb and Waiwitlikhit, Suppakit and Benn, Jason and Kang, Daniel},
  journal = {arXiv preprint arXiv:2506.19733},
  year    = {2025},
  url     = {https://arxiv.org/abs/2506.19733}
}

@inproceedings{li2025dmo,
  title     = {Data Mixing Optimization for Supervised Fine-Tuning of Large Language Models},
  author    = {Li, Yuan and Liu, Zhengzhong and Xing, Eric},
  booktitle = {Proceedings of the 42nd International Conference on Machine Learning (ICML)},
  year      = {2025},
  publisher = {PMLR},
  url       = {https://openreview.net/pdf?id=19kqoNoc2N}
}

@article{zhu2025internvl3,
  title={Internvl3: Exploring advanced training and test-time recipes for open-source multimodal models},
  author={Zhu, Jinguo and Wang, Weiyun and Chen, Zhe and Liu, Zhaoyang and Ye, Shenglong and Gu, Lixin and Tian, Hao and Duan, Yuchen and Su, Weijie and Shao, Jie and others},
  journal={arXiv preprint arXiv:2504.10479},
  year={2025}
}
